\documentclass[lettersize,journal]{IEEEtran}

\usepackage[ruled]{algorithm2e} 
\usepackage{multirow}
\usepackage[table,xcdraw]{xcolor}
\usepackage[normalem]{ulem}
\useunder{\uline}{\ul}{}

\usepackage{amsmath,amsfonts}
\usepackage[caption=false,font=normalsize,labelfont=sf,textfont=sf]{subfig}
\usepackage{verbatim}
\usepackage{graphicx}
\usepackage{cite}

\begin{document}
\title{CrossView-GS: Cross-view Gaussian Splatting For Large-scale Scene Reconstruction}

\author{Chenhao Zhang, Yuanping Cao and Lei Zhang*,~\IEEEmembership{Member,~IEEE,}
\thanks{Chenhao Zhang, Yuanping Cao and Lei Zhang are with the School of Computer Science, Beijing Institute of Technology, Beijing 100081, China.}
\thanks{* Corresponding author: leizhang@bit.edu.cn}
}

\maketitle

\begin{abstract}
3D Gaussian Splatting (3DGS) leverages densely distributed Gaussian primitives for high-quality scene representation and reconstruction. While existing 3DGS methods perform well in scenes with minor view variation, large view changes from cross-view data pose optimization challenges for these methods. To address these issues, we propose a novel cross-view Gaussian Splatting method for large-scale scene reconstruction based on multi-branch construction and fusion. Our method independently reconstructs models from different sets of views as multiple independent branches to establish the baselines of Gaussian distribution, providing reliable priors for cross-view reconstruction during initialization and densification. Specifically, a gradient-aware regularization strategy is introduced to mitigate smoothing issues caused by significant view disparities. Additionally, a unique Gaussian supplementation strategy is utilized to incorporate complementary information of multi-branch into the cross-view model. Extensive experiments on benchmark datasets demonstrate that our method achieves superior performance in novel view synthesis compared to state-of-the-art methods.
\end{abstract}


\begin{figure*}[htbp]
  \includegraphics[width=\textwidth]{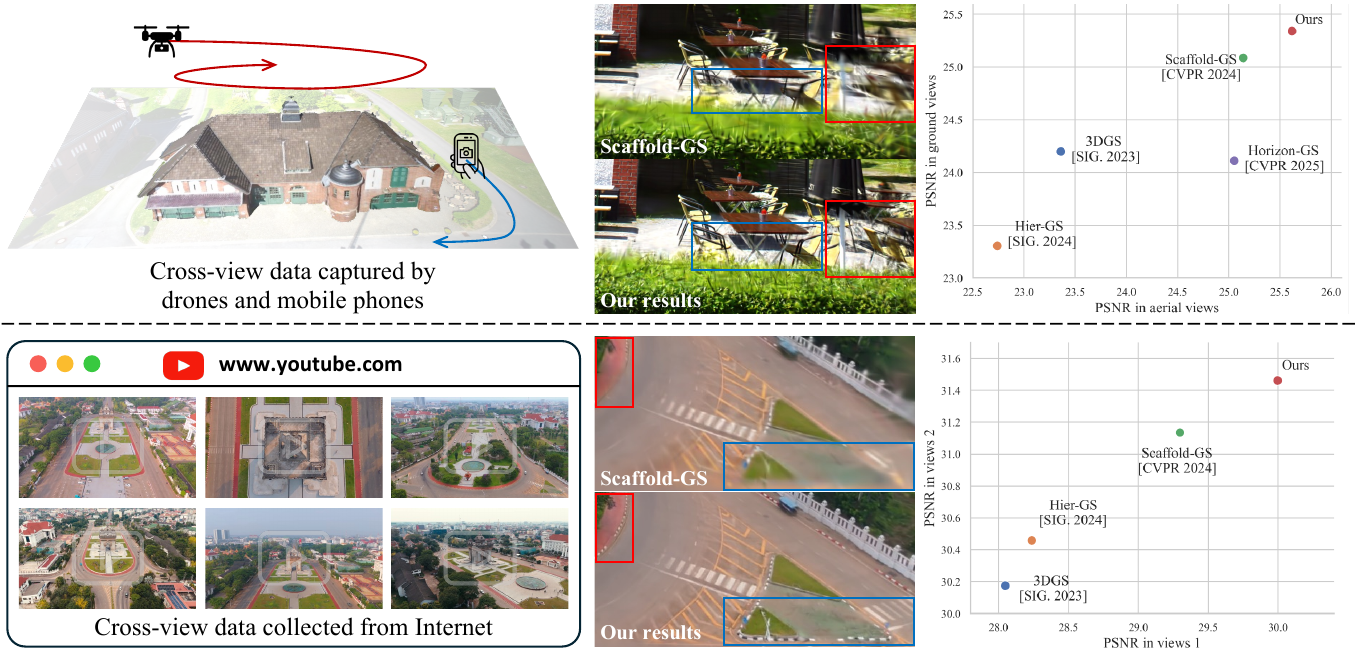}
  \caption{\textbf{Left:} The cross-view data poses challenges for large-scale scene reconstruction due to significant view variation. \textbf{Middle:} Our method CrossView-GS, demonstrates the ability to reconstruct finer structures with more consistent geometry and appearance than Scaffold-GS~\cite{scaffold} across diverse scenes. \textbf{Right:} The quantitative comparisons with additional SOTA methods highlight the superiority of our method in reconstructing scenes using cross-view data.}
  \label{fig:teaser}
\end{figure*}
\section{Introduction}
3D Gaussian Splatting (3DGS) represents 3D scenes through the use of densely distributed Gaussian primitives, which has increasingly emerged as a prominent approach for scene representation and reconstruction~\cite{review1, review2}. Recent advancements in mobile sensors and unmanned devices have made it feasible to capture large-scale scenes at a low cost using mobile phones or drones. However, due to the extensive scale of the scenes, it is often essential to utilize multiple devices or trajectories to gather data from significantly varied viewpoints, thereby generating the so-called cross-view data, rather than one device and trajectory with continuous and uniform viewpoints. A notable example involves capturing a scene from both aerial and ground perspectives by using drones and mobile phones, or data sourced from the Internet (see Fig.~\ref{fig:teaser}). Such cross-view data holds substantial value in virtual reality, smart cities, and geographic information systems, \emph{etc}.

3DGS typically assumes that the input images originate from a set of views with minor variations surrounding the scene. So it can achieve Adaptive Densification Control (ADC) by calculating the average gradient from randomly selected views within fixed intervals~\cite{3dgs, scaffold}. However, significant differences between views lead to challenges with ADC~\cite{mvgs, ucgs}. When reconstructing scenes from cross-view images, salient gradients in a particular view may become excessively smoothed by other views exhibiting large changes, adversely affecting the densification of Gaussian primitives that rely on these salient gradients. This ultimately results in unsatisfied reconstruction outcomes, which may even be inferior to those achieved using data from individual views (see results of Scaffold-GS in Fig.~\ref{fig:teaser}). Some methods like \cite{dragon}, use 3DGS based on aerial views to  extrapolate lower-altitude views for approximating ground views, but they usually rely on the manual pre-specification of camera poses at intermediate altitudes and robustness of  perceptual regularization functions in novel view synthesis. Thus, effectively leveraging complementary information from cross-view data for reconstruction remains insufficiently explored.

Actually, classical methods~\cite{cross-view_geo-loc, CVM-Net, bridging} predominantly focusing on ground-to-aerial images, utilize separate branches to extract features from individual views and address the challenges of significant view differences by fusing the unique information. Inspired by this methodology, we adopt multi-branch models that are constructed separately to establish the baseline Gaussian distributions for views with substantial discrepancies. These distributions serve as reliable priors that guide the optimization of cross-view reconstruction throughout both the initialization and densification processes. By addressing the optimization challenges arising from substantial view variations in cross-view data, this approach enables a comprehensive reconstruction of large-scale scenes (see Fig. \ref{fig:teaser}).

The main contribution of our work is a novel large-scale scene reconstruction method with cross-view data as input. The multiple branches are constructed based on different sets of views as priors, generating point clouds as initialization and followed by gradient-aware regularization in the optimization process for implicit fusion. A unique Gaussians supplementation strategy is also adopted to explicitly fuse complementary information derived from different branches. 
Experimental results demonstrate that the proposed method effectively addresses the optimization challenges of 3DGS in reconstructing large-scale scenes with cross-view data.

\section{Related Work}


\subsection{Reconstruction based on 3DGS}
Recently, several studies have sought to enhance the densification within the 3DGS framework. Pixel-GS~\cite{pixel} promotes the densification of large Gaussian primitives by considering the number of pixels covered by Gaussians in each view. \cite{revising} proposes using a per-pixel error function as the criteria for densification and correcting biases in opacity handling strategies. Although these methods have successfully improved the densification strategy of the original 3DGS, none has addressed the gradient smoothing issue caused by significant view changes.

Several methods based on 3DGS address multi-scale rendering issues. Mip-Splatting~\cite{mipsplatting} grounded in the Nyquist-Shannon sampling theorem, introduces a 3D smoothing filter and a 2D Mip filter to constrain frequency based on the maximal observed sampling rate. BungeeNeRF~\cite{bungeenerf} also demonstrates that as the camera gradually descends from the aerial views, multi-scale optimization issues arise. However, the drastic view changes encountered in cross-view data are independent of multi-scale rendering issues. Such drastic changes violate the assumptions governing ADC in 3DGS, leading to unreasonable gradient statistics. This makes it particularly difficult to effectively identify and adjust the Gaussian primitives corresponding to artifacts.

Some studies based on 3DGS have also focused on large-scale scene reconstruction. CityGaussian~\cite{citygs} proposes a divide-and-conquer training method and a real-time level-of-detail (LoD) rendering strategy for large-scale scene rendering. VastGaussian~\cite{vastgs} through a progressive partitioning strategy and appearance-decoupled modeling, achieves rapid optimization and high-fidelity real-time rendering, while Hier-GS~\cite{hiergs} introduces a hierarchical structure with a LoD scheme for real-time rendering. However, these methods typically assume a single set of views as input, which do not adequately address the complexities of comprehensive scene reconstruction using cross-view data.

\subsection{Reconstruction based on cross-view data}
The concept of cross-view originates from the task of ``cross-view image geo-localization''~\cite{cross-view}, which is limited to cross-view data composed of aerial and ground views, and traditionally based on photogrammetry~\cite{gao2019ground, qin20203d}.

With the advent of 3DGS, UC-GS~\cite{ucgs} proposes leveraging aerial views to assist ground-view reconstruction, thereby improving the performance of autonomous driving. DRAGON~\cite{dragon} performs joint registration and reconstruction by iteratively extrapolating 3DGS from aerial to ground views via manually defined poses, which limits its generalization ability and faces a trade-off between extrapolation extent and rendering quality or efficiency. Some recent methods~\cite{zhu2020leveraging, dust3r} offer more robust cross-view registration for reconstruction. Horizon-GS~\cite{horizongs} focuses on tackling efficiency challenges in large-scale scenes, but fails to effectively address the densification issues  in the context of cross-view data. In contrast, our work attempts to optimize 3DGS in cross-view reconstruction while automating the comprehensive reconstruction of large-scale scenes.

\begin{figure*}[htbp]
  \includegraphics[width=\textwidth]{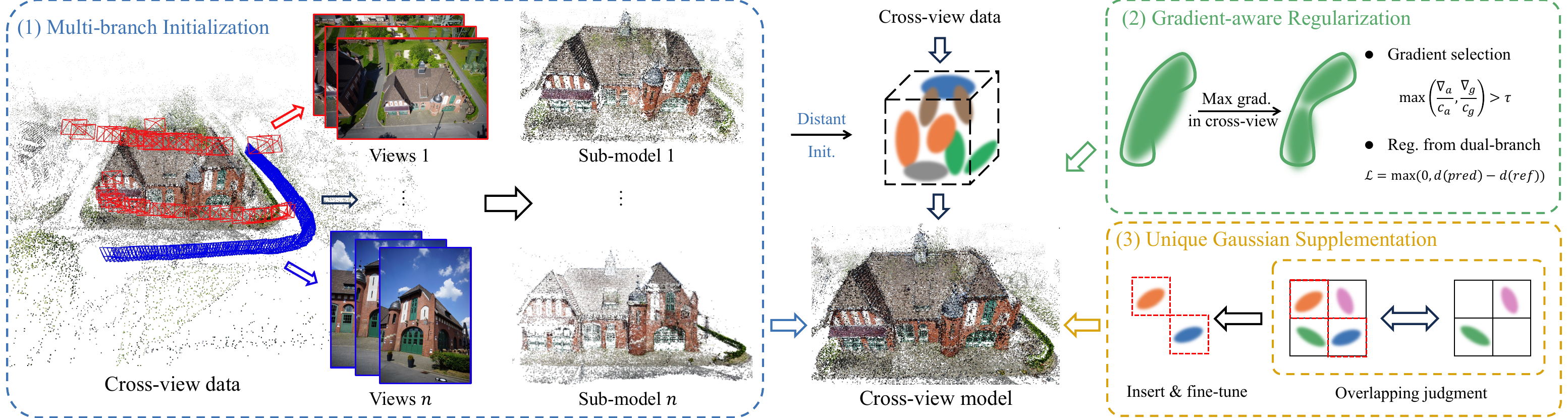}
  \caption{\textbf{An overview of the proposed method CrossView-GS}. 
  The method first initializes the multi-branch structure comprising several sub-models while establishing the cross-view model based on distant views. Then,  a gradient-sensitive regularization strategy is used to reconstruct the cross-view model. Finally, unique Gaussian primitives are fused to supplement the cross-view model, achieving large-scale scene reconstruction through fine-tuning. 
  }
  \label{fig:pipeline}
\end{figure*}

\section{Method Overview}
We outline the proposed method to establish context for the core algorithm described in the next section. As shown in Fig.~\ref{fig:pipeline}, our method seeks to reconstruct large-scale scenes from a diverse set of cross-view images captured by cameras mounted on drones or ground devices along multiple trajectories or angles. 

Specifically, we first train the multi-branch models which containing multiple sub-models independently using a set of views corresponding to views from different trajectories. Structured point clouds extracted from the distant-view sub-model serve as the initial distribution for cross-view reconstruction. Then, the gradient-aware regularization based on pseudo-labels generated by the multi-branch models is applied to optimize the reconstruction process. Finally, by assessing overlapping regions from the multi-branch models, unique Gaussian primitives are integrated into the cross-view reconstruction, with the final model completed through fine-tuning.

\section{Cross-view reconstruction}
We initially establish multiple branches as priors. Subsequently, we enhance the ADC by leveraging the characteristics of gradients in cross views and fully fuse unique information from different views.

\subsection{Multi-branch initialization}

To address the optimization challenges caused by significant view changes, we follow the Siamese-like architecture during the initial stage, using the state-of-the-art 3DGS method Scaffold-GS~\cite{scaffold} to reconstruct scenes from $n$ sets of views $\{v_i\mid i=1,2,\ldots,n\}$, resulting in the sub-models $\{G_i\mid i=1,2,\ldots,n\}$. The set of 3D Gaussian primitives is defined as $\mathcal{G}=\{G_i\left(\mu_i,\Sigma_i,\alpha_i,f_i\right)\mid i=1,2,\ldots,N\}$, where $\mu_i$, $\Sigma_i$, $\alpha_i$, and $f_i$ represent the mean, covariance matrix, transparency, and feature vector (RGB color or spherical harmonics) of the $i$-th Gaussian primitive, respectively.

\textbf{Initialization based on distant views.} Inspired by PyGS~\cite{pygs}, to obtain a more comprehensive structured point cloud for large-scale scenes as the initial distribution, we downsample the Gaussian distribution reconstructed from distant views (such as aerial views in aerial-ground scenes) to generate the point cloud $pc$. This process can be expressed as: 
\begin{equation}
\mathcal{G}\prime=\mathrm{Downsample}\left(\mathcal{G},\tau\right), 
\end{equation}
where $\tau$ is the downsampling ratio. Based on the downsampled Gaussian distribution $\mathcal{G}^\prime$, the center position $\mu_i^\prime$ of each Gaussian primitive $G_i^\prime$ is selected to generate the point cloud as $pc=\cup\mu_i\prime$. Subsequently, we use $pc$ as the initial point cloud to reconstruct the cross-view model $\mathcal{G}_f$ based on cross-view images.

\subsection{Gradient-aware regularization}

Due to significant view differences, 3DGS faces challenges in reconstruction based on cross-view images. Our observations indicate that these challenges mainly stem from the limitations of the densification strategy in ADC. The densification strategy of 3DGS relies on gradient statistics within fixed time steps. Taking Scaffold-GS as an example, for each Gaussian primitive $G$, the cumulative gradient $\nabla_\pi$ and cumulative visibility count $c_\pi$ over $N$ training iterations are calculated based on the training camera poses $\pi$, and the average gradient is then computed as $\nabla_{avg}=\nabla_\pi/c_\pi$. Next, the set of Gaussian primitives satisfying $\nabla_{avg}>\tau$ is selected, where $\tau$ is a pre-defined threshold denoted by $\mathcal{G}_\pi=\left\{G\middle|\frac{\nabla_\pi}{c_\pi}>\tau\right\}$. If the voxel containing the Gaussian primitives $\mathcal{G}_\pi$ does not already have an anchor, a new anchor is deployed at the center of the voxel. 

With views from either $v_1$ or $v_2$, which exhibit significant differences, the densified set of Gaussian primitives is 

\begin{equation}
\left\{G\middle|\frac{\nabla_v}{c_v}>\tau,\ v\in v_1\vee\ v\in v_2\right\}.
\end{equation}

Then, using cross-view data $v_f$ composed of this two views, the densified set of Gaussian primitives is 

\begin{equation}
\left\{G\middle|\frac{\nabla_f}{c_f}>\tau,f=v_1\cup v_2\right\}=\left\{G\middle|\frac{\nabla_1+\nabla_2}{c_1+c_2}>\tau\right\}. 
\end{equation}

\begin{figure}[tbp]
  \includegraphics[width=\linewidth]{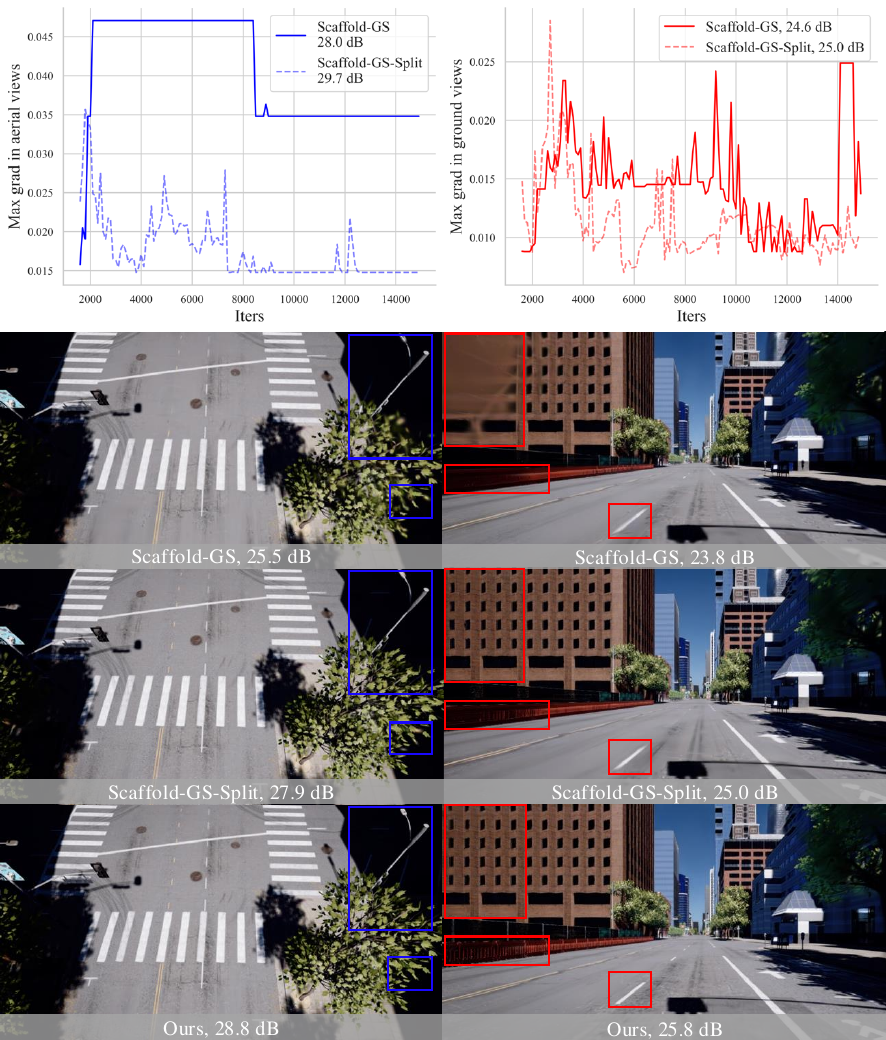}
  \caption{
  The first row shows the variation of maximum gradient in different views during densification when using cross-view or single-view for reconstruction. The second and third rows show the results of Scaffold-GS~\cite{scaffold} using cross-view data and single-view data, respectively. The forth row shows the result of our CrossView-GS using cross-view data.}
  \label{fig:grad_smooth}
\end{figure}

Since the max values of $\frac{\nabla_1+\nabla_2}{c_1+c_2}$, $\frac{\nabla_1}{c_1}$ and $\frac{\nabla_2}{c_2}$ are only equal when $\frac{\nabla_1}{c_1}=\frac{\nabla_2}{c_2}$, the salient gradients for densification are prone to being smoothed out. This leads the model to fall into local optima, preventing it from achieving the same densification effect as in single-view reconstruction (see Fig.~\ref{fig:grad_smooth}).

During the training of the cross-view model $\mathcal{G}_f$, we propose a gradient-aware regularization to alleviate the problem of salient gradients being smoothed out. This regularization strategy consists of two components: an adjustment to the gradient selection criterion for cross-view images and the additional regularization constraints based on the multi-branch models.

Firstly, large view changes may cause inconsistencies in the gradients of Gaussian primitives across different views. To address this issue, we modify the gradient selection criterion in the densification strategy to accommodate cross-view variations. Specifically, to accurately identify Gaussian primitives with salient gradients based on cross-view images and achieve densification effects comparable to single-view construction, we separately record the gradients ($\nabla_1$ and $\nabla_2$) and visibility counts ($c_1$ and $c_2$) of Gaussian primitives under different sets of views. The densified Gaussian primitive set $\mathcal{G}_{cross}$ is then determined using the maximum gradient from the single type of views:

\begin{equation}
\mathcal{G}_{cross}=\left\{G\middle|max(\frac{\nabla_1}{c_1},\frac{\nabla_2}{c_2})>\tau\right\}
\end{equation}

Next, inspired by the triplet loss~\cite{facenet}, we use the rendering of the trained multi-branch models under the corresponding views as the pseudo label of regularization. By penalizing regions in the cross-view model where the reconstruction quality is lower than that of the multi-branch models, additional gradients are introduced. Specifically, given a training view $v_t$, the rendering of the multi-branch models $\mathcal{G}_1$ and $\mathcal{G}_2$ are used as the pseudo label $ref$. An additional loss term penalizes parts of the cross-view model’s rendering $pred$ that deviate further from the ground truth $gt$ than pseudo label $ref$. This process can be expressed as:

\begin{equation}
\mathcal{L}_{reg}=\max{\left(0,\ d\left(pred,gt\right)-d\left(ref,gt\right)\right)},
\end{equation}
where $d(\cdot, \cdot)$ is a distance metric function, and $d\left(ref,gt\right)$ is the reference distance. This ensures that $pred$ is optimized to be closer to $gt$ than $ref$, while the $\max()$ function ensures that this loss term is effective only when $pred$ deviates further from $gt$ than $ref$.

\begin{table*}[tb]
\caption{Quantitative evaluation of our method CrossView-GS and SOTA methods with Block\_small scene in MatrixCity~\cite{mc}, Zeche scene in ISPRS~\cite{isprs}, NYC and SF scenes in UC-GS~\cite{ucgs}. The \textbf{best} and {\ul second-best} ones are highlighted in the table.}
\label{tab:quality}
\centering
\begin{tabular}{cc|ccccc|ccccc}
\hline
\multicolumn{2}{c|}{\multirow{2}{*}{Scene}} &
  \multicolumn{5}{c|}{Block\_small} &
  \multicolumn{5}{c}{Zeche} \\ \cline{3-12} 
\multicolumn{2}{c|}{} &
  \multicolumn{3}{c|}{All} &
  \multicolumn{1}{c|}{Aerial} &
  Ground &
  \multicolumn{3}{c|}{All} &
  \multicolumn{1}{c|}{Aerial} &
  Ground \\ \hline
\multicolumn{1}{c|}{Methods} &
  Metrics &
  PSNR ↑ &
  SSIM ↑ &
  \multicolumn{1}{c|}{LPIPS ↓} &
  \multicolumn{1}{c|}{PSNR ↑} &
  PSNR ↑ &
  PSNR ↑ &
  SSIM ↑ &
  \multicolumn{1}{c|}{LPIPS ↓} &
  \multicolumn{1}{c|}{PSNR ↑} &
  PSNR ↑ \\ \hline
3DGS &
  SIG. 2023 &
  24.121 &
  0.810 &
  \multicolumn{1}{c|}{0.284} &
  \multicolumn{1}{c|}{25.413} &
  23.777 &
  23.822 &
  0.779 &
  \multicolumn{1}{c|}{0.251} &
  \multicolumn{1}{c|}{23.361} &
  24.198 \\
Mip-Splatting &
  CVPR 2024 &
  22.830 &
  0.645 &
  \multicolumn{1}{c|}{0.553} &
  \multicolumn{1}{c|}{21.134} &
  23.282 &
  23.099 &
  0.771 &
  \multicolumn{1}{c|}{0.265} &
  \multicolumn{1}{c|}{23.090} &
  23.107 \\
Scaffold-GS &
  CVPR 2024 &
  24.596 &
  {\ul 0.836} &
  \multicolumn{1}{c|}{0.228} &
  \multicolumn{1}{c|}{{\ul 27.959}} &
  23.699 &
  {\ul 25.111} &
  0.810 &
  \multicolumn{1}{c|}{0.227} &
  \multicolumn{1}{c|}{{\ul 25.142}} &
  {\ul 25.085} \\
Hier-GS &
  SIG. 2024 &
  24.410 &
  0.813 &
  \multicolumn{1}{c|}{0.267} &
  \multicolumn{1}{c|}{26.381} &
  23.885 &
  23.050 &
  0.748 &
  \multicolumn{1}{c|}{0.279} &
  \multicolumn{1}{c|}{22.740} &
  23.304 \\
Horizon-GS &
  CVPR 2025 &
  {\ul 24.776} &
  \textbf{0.846} &
  \multicolumn{1}{c|}{\textbf{0.197}} &
  \multicolumn{1}{c|}{27.800} &
  {\ul 23.970} &
  24.535 &
  {\ul 0.813} &
  \multicolumn{1}{c|}{{\ul 0.215}} &
  \multicolumn{1}{c|}{25.054} &
  24.110 \\
\multicolumn{2}{c|}{CrossView-GS (Ours)} &
  \textbf{25.233} &
  \textbf{0.846} &
  \multicolumn{1}{c|}{{\ul 0.213}} &
  \multicolumn{1}{c|}{\textbf{28.464}} &
  \textbf{24.371} &
  \textbf{25.465} &
  \textbf{0.824} &
  \multicolumn{1}{c|}{\textbf{0.201}} &
  \multicolumn{1}{c|}{\textbf{25.618}} &
  \textbf{25.339} \\ \hline
\multicolumn{2}{c|}{\multirow{2}{*}{Scene}} &
  \multicolumn{5}{c|}{NYC} &
  \multicolumn{5}{c}{SF} \\ \cline{3-12} 
\multicolumn{2}{c|}{} &
  \multicolumn{3}{c|}{All} &
  \multicolumn{1}{c|}{Aerial} &
  Ground &
  \multicolumn{3}{c|}{All} &
  \multicolumn{1}{c|}{Aerial} &
  Ground \\ \hline
\multicolumn{1}{c|}{Methods} &
  Metrics &
  PSNR ↑ &
  SSIM ↑ &
  \multicolumn{1}{c|}{LPIPS ↓} &
  \multicolumn{1}{c|}{PSNR ↑} &
  PSNR ↑ &
  PSNR ↑ &
  SSIM ↑ &
  \multicolumn{1}{c|}{LPIPS ↓} &
  \multicolumn{1}{c|}{PSNR ↑} &
  PSNR ↑ \\ \hline
3DGS &
  SIG. 2023 &
  26.574 &
  0.814 &
  \multicolumn{1}{c|}{0.249} &
  \multicolumn{1}{c|}{27.654} &
  25.494 &
  26.705 &
  0.740 &
  \multicolumn{1}{c|}{0.347} &
  \multicolumn{1}{c|}{28.483} &
  24.928 \\
Mip-Splatting &
  CVPR 2024 &
  24.273 &
  0.769 &
  \multicolumn{1}{c|}{0.290} &
  \multicolumn{1}{c|}{25.006} &
  23.540 &
  26.006 &
  0.739 &
  \multicolumn{1}{c|}{0.339} &
  \multicolumn{1}{c|}{27.444} &
  24.569 \\
Scaffold-GS &
  CVPR 2024 &
  27.255 &
  0.825 &
  \multicolumn{1}{c|}{0.232} &
  \multicolumn{1}{c|}{28.730} &
  25.780 &
  26.817 &
  0.751 &
  \multicolumn{1}{c|}{0.324} &
  \multicolumn{1}{c|}{28.754} &
  24.879 \\
Hier-GS &
  SIG. 2024 &
  {\ul 27.617} &
  0.837 &
  \multicolumn{1}{c|}{0.200} &
  \multicolumn{1}{c|}{29.157} &
  {\ul 26.076} &
  27.633 &
  {\ul 0.784} &
  \multicolumn{1}{c|}{0.281} &
  \multicolumn{1}{c|}{29.635} &
  \textbf{25.631} \\
Horizon-GS &
  CVPR 2025 &
  27.554 &
  {\ul 0.839} &
  \multicolumn{1}{c|}{\textbf{0.189}} &
  \multicolumn{1}{c|}{\textbf{30.042}} &
  25.067 &
  {\ul 27.663} &
  \textbf{0.791} &
  \multicolumn{1}{c|}{\textbf{0.240}} &
  \multicolumn{1}{c|}{{\ul 29.948}} &
  {\ul 25.379} \\
\multicolumn{2}{c|}{CrossView-GS (Ours)} &
  \textbf{28.116} &
  \textbf{0.843} &
  \multicolumn{1}{c|}{{\ul 0.199}} &
  \multicolumn{1}{c|}{{\ul 29.860}} &
  \textbf{26.373} &
  \textbf{27.702} &
  {\ul 0.784} &
  \multicolumn{1}{c|}{{\ul 0.269}} &
  \multicolumn{1}{c|}{\textbf{30.029}} &
  25.374 \\ \hline
\end{tabular}
\end{table*}

\subsection{Unique Gaussian supplementation}

To fully integrate complementary information from cross-view images, we determine whether Gaussian primitives in the single sub-models and the cross-view model overlap based on a voxel grid. This strategy allows us to identify the densified regions unique to each branch during their respective training processes. Subsequently, these unique Gaussian primitives are incorporated into the cross-view model and further optimized through fine-tuning. Specifically, the multi-branch models ($\mathcal{G}_1$ or $\mathcal{G}_2$), and the cross-view model $\mathcal{G}_f$ are reconstructed using Scaffold-GS with the same voxel size. The voxel grid represents the valid mask of anchors. Define the voxel grid of the cross-view model as $\mathcal{V}_f$ and those of the multi-branch models as $\mathcal{V}_1$ and $\mathcal{V}_2$. A Gaussian primitive $G_i\in\mathcal{G}_1\cup\mathcal{G}_2$ is considered to be non-overlapping with the cross-view model if its anchor’s corresponding voxel $\nu_i$ satisfies $\nu_i\notin\mathcal{V}_f$. Finally, all non-overlapping Gaussian primitives from the multi-branch models are incorporated into the cross-view model. Here, the updated cross-view model $\mathcal{G}_f\prime$ is expressed as:
\begin{equation}
\mathcal{G}_f\prime=\mathcal{G}_f\cup\{G_i\mid G_i\in\mathcal{G}_1\cup\mathcal{G}_2,\nu_i\notin\mathcal{V}_f\}.
\end{equation}

Considering that neural Gaussians in Scaffold-GS possess implicit attributes related to MLPs (e.g., color and opacity), a small amount of iterations of fine-tuning are required for the updated cross-view model $\mathcal{G}_f\prime$. This ensures that the supplemented Gaussian primitives correctly decode appearance information (see detailed analysis in Sec. B in the supplementary document).

\textbf{Loss design.} We basically follow the design of Scaffold-GS~\cite{scaffold}, using the loss function $L_s$ composed of per-pixel color loss $L_1$, SSIM term~\cite{ssim} $L_{SSIM}$, and volume regularization~\cite{volume_reg} $L_{vol}$. In the fine-tune phase, we use the loss function $L_s$, which is defined by:
\begin{equation}
    \mathcal{L}_{s} = \mathcal{L}_{1} + \lambda_{SSIM}\mathcal{L}_{SSIM} + \lambda_{vol}\mathcal{L}_{vol},
\end{equation}

In addition, we also introduce the gradient-aware regularization term $L_{reg}$ during the first training of cross-view model to penalize regions with lower rendering quality than the multi-branch models. The total loss function $L$ is expressed as:

\begin{equation}
\mathcal{L}=\lambda_{reg}\mathcal{L}_{reg}+\mathcal{L}_{s},
\end{equation}

\section{Experiments}

\textbf{Baselines.} We compare our method with state-of-the-art (SOTA) methods, like 3DGS~\cite{3dgs}, Mip-Splatting~\cite{mipsplatting}, Scaffold-GS~\cite{scaffold}, Hier-GS~\cite{hiergs} and Horizon-GS~\cite{horizongs}. For 3DGS and Scaffold-GS, to ensure a fair comparison, we adopt the same number of training iterations (50k) as used in our cross-view model. For Horizon-GS with 60k and 40k iterations for coarse and fine training respectively, we use half of these iterations to maintain fairness. For Hier-GS, we use the default 100k training iterations and record the rendering quality of its leaves. All GS-based methods are initialized with the same point cloud.

\textbf{Datasets.} As done by the SOTA methods, we use Block\_small and Block\_A in MatrixCity dataset~\cite{mc}, NYC and SF in UC-GS dataset~\cite{ucgs} and real captured Zeche scene in the ISPRS dataset~\cite{isprs} for evaluation. For the UC-GS dataset, we utilize the scene captured at an aerial altitude of 20 meters and a ground altitude of 1.5 meters. To better evaluate the reconstruction quality of cross-view scenes, the dataset is re-divided into training and test sets by selecting one test image out of every eight, based on the image suffix number. In addition, we also conduct quantitative experiments with the original settings of the UC-GS dataset. 

To demonstrate the generalization ability of our method, we collect publicly available videos casually captured at two landmark locations: Wat Nong Waeng Temple and Patuxai Monument, sourced from the Internet. Subsequently, we construct two datasets by performing simple trajectory-based clustering with several sets of views, named Temple and Monument respectively. We provide more details in Sec. A in Supplementary Material.

\begin{figure*}[tbp]
  \centering
  \includegraphics[width=0.98\textwidth]{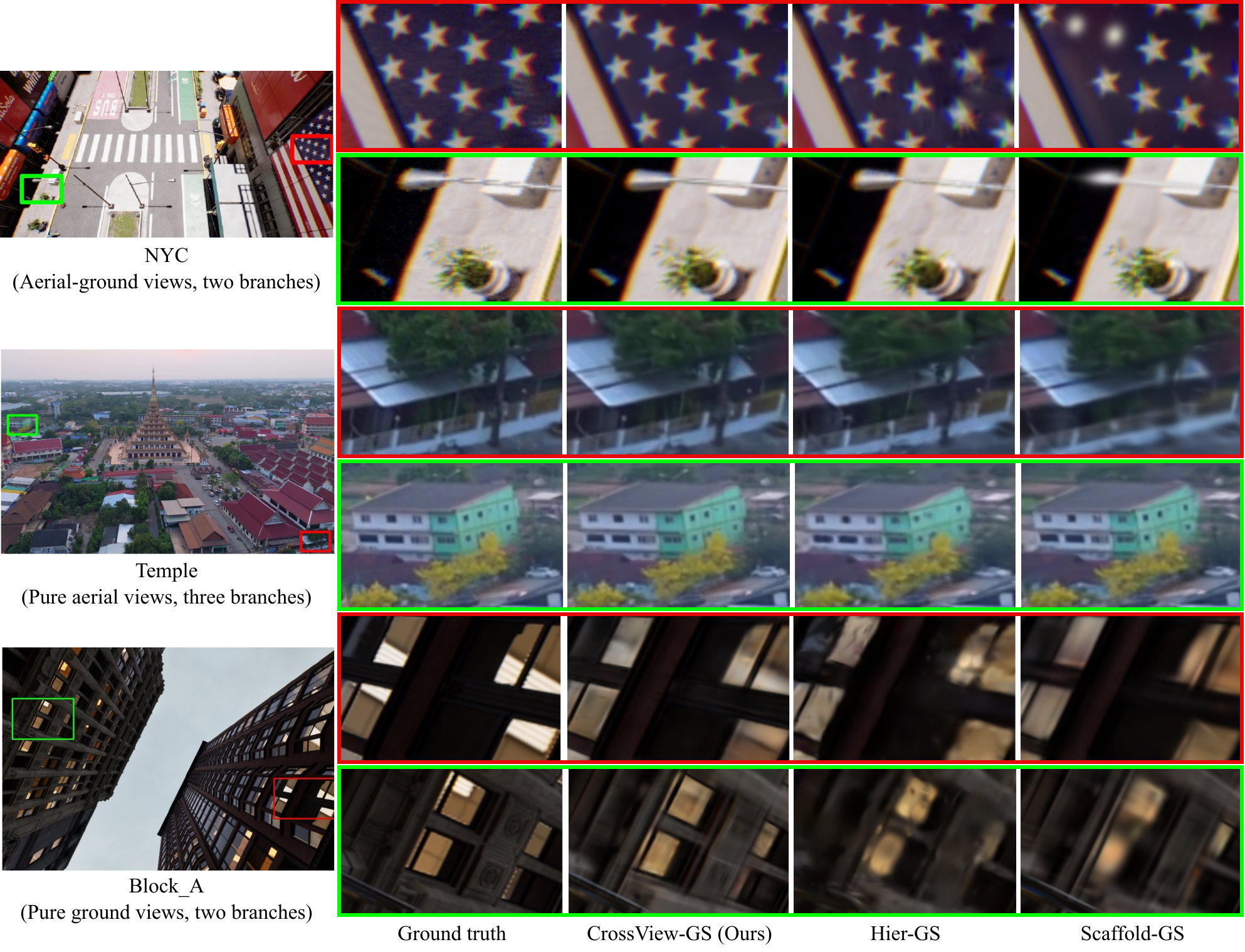}
  \caption{The rendering results and zoomed-in images obtained by using our CrossView-GS, as well as some SOTA methods like Hier-GS and Scaffold-GS.}
  \label{fig:quality}
\end{figure*}


\textbf{Parameter setting.} We basically follow the settings of Scaffold-GS. We train the multi-branch models and the cross-view model with gradient-aware regularization through 30k iterations. Then, we implement unique Gaussian supplementation to cross-view model and finetune it for 20k iterations. The downscale rate $\tau$ is set to 10. For the loss function, parameters $\lambda_{SSIM}$, $\lambda_{vol}$ and $\lambda_{reg}$ are set to 0.2, 0.01 and 1.0. Experiments are done with NVIDIA RTX A6000 GPUs. 

\subsection{Comparisons}
We have conducted quantitative and qualitative evaluations of view synthesis results obtained from our method and SOTA methods across various cross-view data, including the aerial-ground views, pure aerial views and pure ground views. The quantitative evaluation uses the metrics of PSNR, SSIM~\cite{ssim} and LPIPS~\cite{lpips} to assess rendering quality.

\textbf{Aerial-ground views.} The quantitative results of utilizing both aerial and ground views are presented in Tab.~\ref{tab:quality}. Compared to SOTA methods, CrossView-GS demonstrates superior rendering quality in most scenes. Specifically, compared to our baseline Scaffold-GS, CrossView-GS achieves consistent improvements in all scenes, with an average increase of 1.15 dB in PSNR. Notably, the quality improvement in aerial views is more pronounced than in ground views, because aerial views are more affected by gradient smoothing issues, as illustrated in Fig.~\ref{fig:grad_smooth}. Mip-Splatting yields results similar to 3DGS, indicating that the problem of cross-view reconstruction is not related to the sampling frequency. Hier-GS achieves exceptional quality in detail-rich ground views, owing to its level-of-detail structure with strong detail-capturing capability. In contrast, our method gains competitive performance in ground views across other scenes. Since the difference between the test views and training views in the NYC and SF scenarios are relatively small, the methods like Horizon-GS and Hier-GS that have undergone 100k training iterations, can achieve good evaluation results. In the test views of Block\_small, random rotation is added. The performance of our method that has only undergone 50k training iterations surpasses other SOTA methods, demonstrating better generalization ability. 

\begin{figure*}[tbp]
  \includegraphics[width=\textwidth]{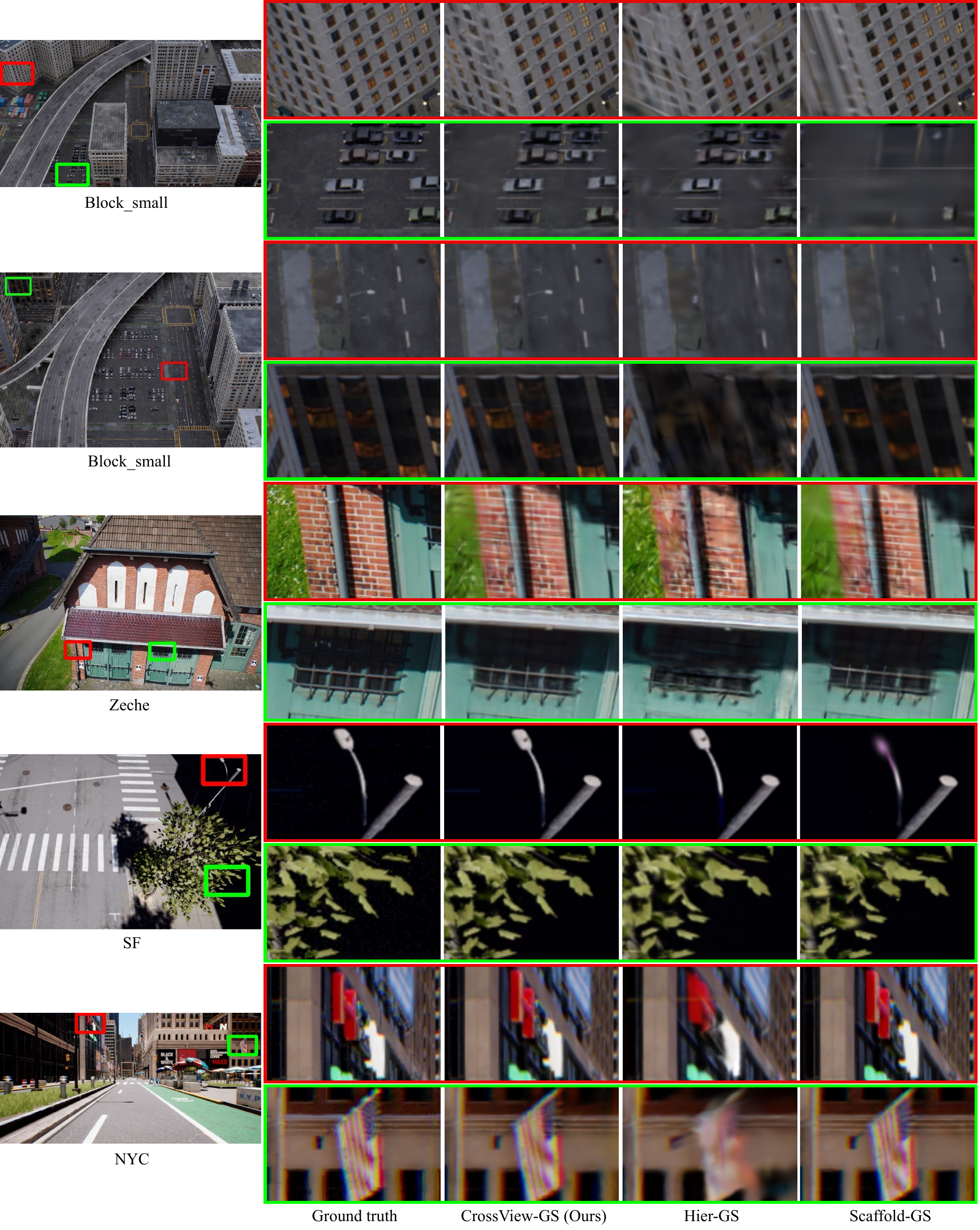}
  \caption{More rendering results and zoomed-in images obtained by our method, as well as some SOTA methods in aerial-ground scenes.}
  \label{fig:quality_appendix}
\end{figure*}

Qualitative comparisons with SOTA methods like Scaffold-GS and Hier-GS are shown in Fig.~\ref{fig:quality} and Fig.~\ref{fig:quality_appendix}, where two views are selected for local magnification to emphasize differences. Scaffold-GS suffers from significant blurring caused by insufficient densification, particularly in aerial views. While Hier-GS exhibits high-quality rendering in ground views, its performance in aerial views is not good. In contrast, our method provides more comprehensive rendering results across both aerial and ground views.

\begin{table}[]
\caption{Quantitative comparison based on Temple and Monument with aerial views and Block\_A with ground views.}
\label{tab:aerial_and_ground}
\setlength{\tabcolsep}{1.4mm} 
\centering
\begin{tabular}{c|ccccc}
\hline
\multirow{3}{*}{Scene} & \multicolumn{5}{c}{Temple \& Monument (aerial views)}                                                  \\ \cline{2-6} 
                       & \multicolumn{3}{c|}{All}                       & \multicolumn{1}{c|}{Views 1} & Views 2 \\ \cline{2-6} 
                       & PSNR ↑ & SSIM ↑ & \multicolumn{1}{c|}{LPIPS ↓} & \multicolumn{1}{c|}{PSNR ↑}  & PSNR ↑  \\ \hline
3DGS                   & 31.620 & 0.925  & \multicolumn{1}{c|}{0.103}   & \multicolumn{1}{c|}{31.451}  & 31.790  \\
Hier-GS                & 31.298 & 0.919  & \multicolumn{1}{c|}{0.114}   & \multicolumn{1}{c|}{30.934}  & 31.774  \\
Scaffold-GS  & 32.968 & 0.940 & \multicolumn{1}{c|}{0.082} & \multicolumn{1}{c|}{32.863} & 33.044 \\
CrossView-GS & \textbf{33.203} & \textbf{0.943} & \multicolumn{1}{c|}{\textbf{0.076}} & \multicolumn{1}{c|}{\textbf{33.088}} & \textbf{33.274} \\ \hline
\multirow{3}{*}{Scene} & \multicolumn{5}{c}{Block\_A (ground views)}                                                            \\ \cline{2-6} 
                       & \multicolumn{3}{c|}{All}                       & \multicolumn{1}{c|}{Views 1} & Views 2 \\ \cline{2-6} 
                       & PSNR ↑ & SSIM ↑ & \multicolumn{1}{c|}{LPIPS ↓} & \multicolumn{1}{c|}{PSNR ↑}  & PSNR ↑  \\ \hline
3DGS                   & 22.192 & 0.768  & \multicolumn{1}{c|}{0.340}   & \multicolumn{1}{c|}{20.723}  & 25.129  \\
Hier-GS                & 22.165 & 0.763  & \multicolumn{1}{c|}{0.353}        & \multicolumn{1}{c|}{20.489}        & 25.517 \\
Scaffold-GS  & 23.106 & 0.789 & \multicolumn{1}{c|}{0.312} & \multicolumn{1}{c|}{22.026} & 26.247 \\
CrossView-GS & \textbf{23.857} & \textbf{0.810} & \multicolumn{1}{c|}{\textbf{0.283}} & \multicolumn{1}{c|}{\textbf{22.195}} & \textbf{27.181} \\ \hline
\end{tabular}
\vspace{-0.3cm}
\end{table}

\begin{figure*}[tbp]
  \includegraphics[width=\textwidth]{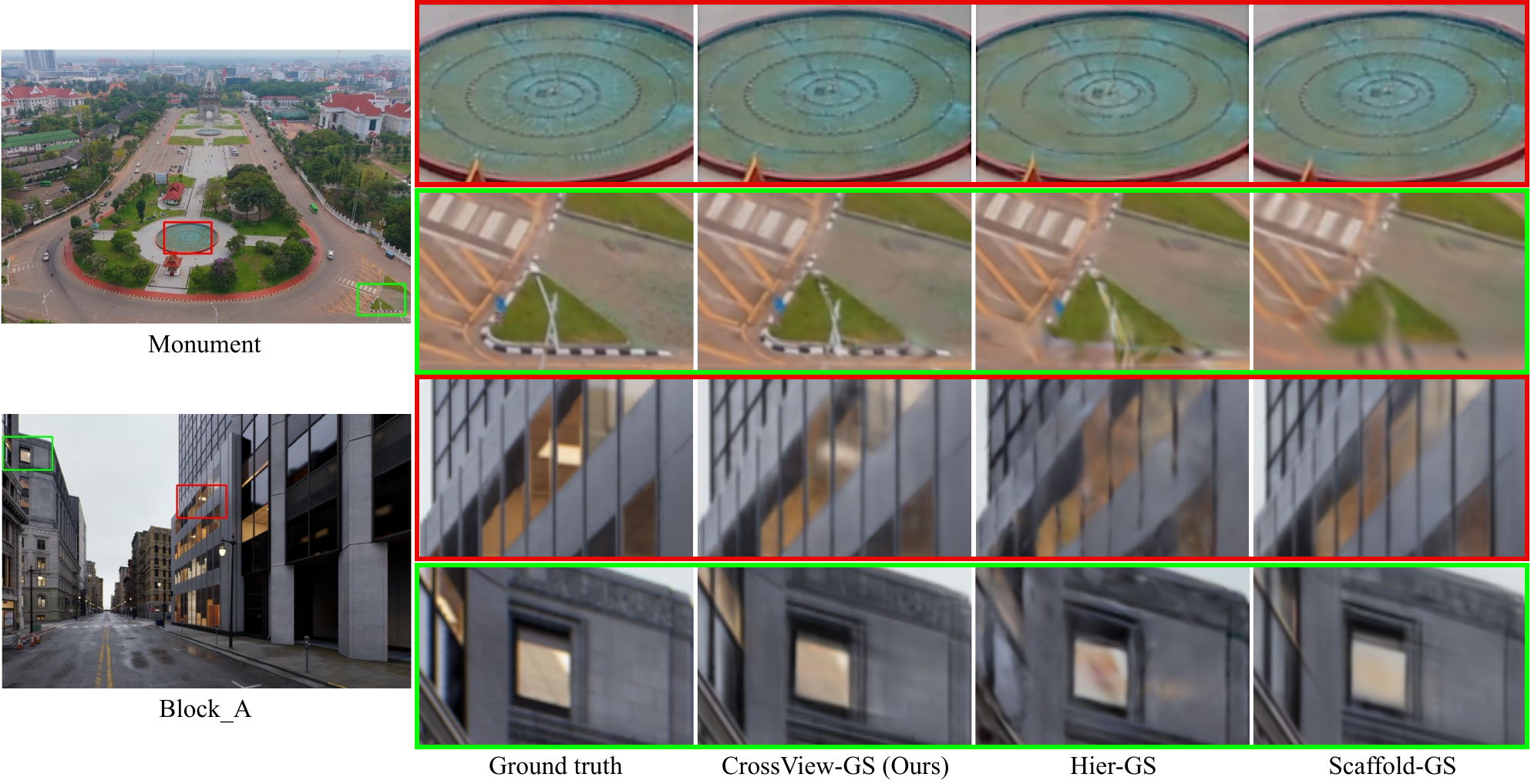}
  \caption{More rendering results and zoomed-in images obtained by our method, as well as some SOTA methods in Monument with aerial views and Block\_A with ground views.}
  \label{fig:quality_appendix_2}
\end{figure*}

\textbf{Pure aerial views.} We also conduct experiments on pure aerial views that exhibit significant differences in the Temple and Monument scenes. In each scene, two sets of views with substantial variations are used as the cross-view data. The average quantitative results are presented in Tab.~\ref{tab:aerial_and_ground}. Although the SOTA methods have shown good rendering quality, our method still achieves the best results on the two sets of views. We also present the qualitative comparison in the first two rows of Fig.~\ref{fig:quality_appendix_2}, where our method successfully improves the quality in details. We further evaluate our method on cross-view data composed of three sets of views in the Temple scene, with the corresponding results shown in Table~\ref{tab:triple}. It is evident that our method can be successfully applied to cross-view data comprising multiple view sets.

\begin{table}[]
\caption{Quantitative comparison based on Temple with aerial views.}
\label{tab:triple}
\setlength{\tabcolsep}{1.0mm} 
\centering
\begin{tabular}{c|cccccc}
\hline
\multirow{3}{*}{Scene} & \multicolumn{6}{c}{Temple}                                                                                             \\ \cline{2-7} 
                       & \multicolumn{3}{c|}{All}                       & \multicolumn{1}{c|}{Views 1} & \multicolumn{1}{c|}{Views 2} & Views 3 \\ \cline{2-7} 
                       & PSNR ↑ & SSIM ↑ & \multicolumn{1}{c|}{LPIPS ↓} & \multicolumn{1}{c|}{PSNR ↑}  & \multicolumn{1}{c|}{PSNR ↑}  & PSNR ↑  \\ \hline
3DGS                   & 34.949 & 0.944  & \multicolumn{1}{c|}{0.081}   & \multicolumn{1}{c|}{33.569}  & \multicolumn{1}{c|}{32.734}  & 37.471  \\
Hier-GS                & 34.941  & 0.943   & \multicolumn{1}{c|}{0.086}        & \multicolumn{1}{c|}{33.550}        & \multicolumn{1}{c|}{33.085}        & 37.289        \\
Scaffold-GS  & 36.275 & 0.958 & \multicolumn{1}{c|}{0.062} & \multicolumn{1}{c|}{35.000} & \multicolumn{1}{c|}{34.147} & 38.649 \\
CrossView-GS & \textbf{36.593} & \textbf{0.960} & \multicolumn{1}{c|}{\textbf{0.057}} & \multicolumn{1}{c|}{\textbf{35.254}} & \multicolumn{1}{c|}{\textbf{34.546}} & \textbf{38.989} \\ \hline  	 	 
\end{tabular}
\end{table}

\textbf{Pure ground views.} We also conduct experiments on pure ground views with significant differences. In the MatrixCity, ground views are captured from five directions. We treat front-back and top views as distinct sets due to their notable differences. More details are reported in Sec. A in Supplementary Material. Quantitative results are presented in Tab.~\ref{tab:aerial_and_ground}, where views 1 and views 2 represent the front-back views and top views in Block\_A, respectively. Although the content in the top views is relatively simple, their significant viewpoint differences from surrounding views still pose challenges for cross-view reconstruction. Compared with SOTA methods, our CrossView-GS significantly improves the rendering quality of the top and front views and recovers more detailed information, as shown in the last row of Fig~\ref{fig:quality} and \ref{fig:quality_appendix_2}.

\subsection{Comparison based on UC-GS dataset}
It is noteworthy that the UC-GS dataset is also reconstructed using cross-view data, specifically aimed at enhancing the performance of ground views. To verify the effectiveness of our method, we compare CrossView-GS with SOTA methods on the UC-GS dataset under its specified settings. The evaluation results for the SOTA methods are derived from the UC-GS paper~\cite{ucgs}, where all models are trained for 900k iterations. In contrast, the training and fine-tuning of the cross-view model, only require 30k and 20k iterations respectively, which is significantly less than the 900k iterations required for UC-GS. Our method demonstrates superior performance in enhancing aerial views within cross-view scenes, whereas Tab.~\ref{tab:ucgs_setting} indicates that our method still achieves competitive results for ground views. As the held-out views are closer to the training views, the longer training iterations for UC-GS yield slightly improved outcomes. However, when performing generalization tests based on the two sets of views with translations and rotations, our method still exhibits better performance in metrics such as PSNR.

\begin{table}[tb]
\caption{Quantitative comparison based on the UC-GS dataset.}
\label{tab:ucgs_setting}
\setlength{\tabcolsep}{0.6mm} 
\centering
\begin{tabular}{cl|ccccccccc}
\hline
\multicolumn{2}{c|}{\multirow{3}{*}{Scene}} &
  \multicolumn{9}{c}{UC-GS} \\ \cline{3-11} 
\multicolumn{2}{c|}{} &
  \multicolumn{3}{c|}{Held-out} &
  \multicolumn{3}{c|}{+0.1m} &
  \multicolumn{3}{c}{+0.1m \& 5°down} \\ \cline{3-11} 
\multicolumn{2}{c|}{} &
  PSNR &
  SSIM &
  \multicolumn{1}{c|}{LPIPS} &
  PSNR &
  SSIM &
  \multicolumn{1}{c|}{LPIPS} &
  PSNR &
  SSIM &
  LPIPS \\ \hline
\multicolumn{2}{c|}{3DGS} &
  23.47 &
  0.668 &
  \multicolumn{1}{c|}{0.406} &
  21.25 &
  0.643 &
  \multicolumn{1}{c|}{0.402} &
  20.83 &
  0.605 &
  0.440 \\ \hline
\multicolumn{2}{c|}{Scaffold-GS} &
  25.32 &
  0.737 &
  \multicolumn{1}{c|}{0.328} &
  23.60 &
  0.714 &
  \multicolumn{1}{c|}{0.335} &
  23.04 &
  0.678 &
  0.375 \\ \hline
\multicolumn{2}{c|}{UC-GS} &
  \textbf{25.95} &
  \textbf{0.763} &
  \multicolumn{1}{c|}{0.291} &
  {\ul 24.15} &
  {\ul 0.741} &
  \multicolumn{1}{c|}{{\ul 0.298}} &
  {\ul 23.52} &
  0.702 &
  0.340 \\ \hline
\multicolumn{2}{c|}{Horizon-GS} &
  24.26 &
  0.733 &
  \multicolumn{1}{c|}{{\ul 0.284}} &
  22.93 &
  0.710 &
  \multicolumn{1}{c|}{0.300} &
  23.78 &
  \textbf{0.724} &
  \textbf{0.286} \\ \hline
\multicolumn{2}{c|}{Ours} &
  {\ul 25.73} &
  {\ul 0.762} &
  \multicolumn{1}{c|}{\textbf{0.281}} &
  \textbf{24.52} &
  \textbf{0.748} &
  \multicolumn{1}{c|}{\textbf{0.285}} &
  \textbf{24.20} &
  {\ul 0.717} &
  {\ul 0.322} \\ \hline
\end{tabular}
\end{table}

\subsection{Ablation study}
\begin{figure}[t]
  \includegraphics[width=\linewidth]{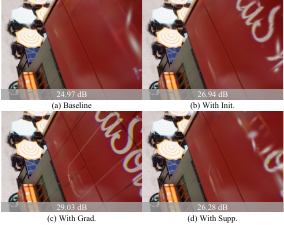}
  \caption{Qualitative ablations. (a) Cross-view reconstruction using the baseline method. (b) Baseline with initialization from distant aerial views. (c) Baseline with gradient-aware regularization. (d) Baseline with unique Gaussian supplementation.}
  \label{fig:ablation_aerial}
\end{figure}

\begin{figure*}[]
  \includegraphics[width=\textwidth]{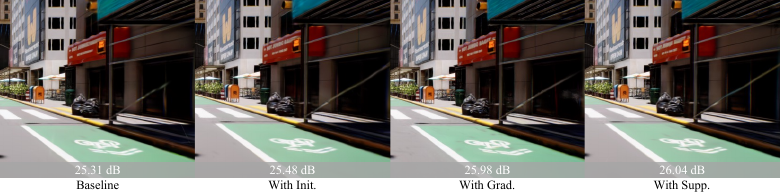}
  \caption{Qualitative ablations. \textbf{Column 1:} The results of baseline in cross-view reconstruction. \textbf{Column 2:} The results of baseline with initialization from distant aerial views. \textbf{Column 3:} The results of baseline with gradient-aware regularization. \textbf{Column 4:} The results of baseline with unique Gaussian supplementation.}
  \label{fig:ablation_street}
\end{figure*}

We conduct ablation experiments on three components of our method using UC-GS dataset~\cite{ucgs}, as shown in Tab.~\ref{tab:ablation}. The evaluation is based on the quantitative metrics like PSNR, SSIM, and LPIPS. Additionally, we select two views with identical horizontal projection position but varying heights and angles to exemplify the ablation of different components, as shown in Fig.~\ref{fig:ablation_aerial} and \ref{fig:ablation_street}.

We first conduct experiments by adding various components to the baseline Scaffold-GS~\cite{scaffold} to demonstrate their effectiveness, as shown in the first four rows in Tab.~\ref{tab:ablation}. It can be seen that the gradient-aware regularization can achieve the maximum performance gain. The point cloud derived from aerial views provides essential structural information, thereby improving the quality of aerial views, as shown in the second column of Fig.~\ref{fig:ablation_aerial}. Furthermore, as shown in the third column, it can be seen that the gradient-aware regularization strategy effectively solves the gradient smoothing issue shown in Fig.~\ref{fig:grad_smooth}, leading to a substantial improvement in the rendering quality of aerial views. In contrast, the unique Gaussian supplementation enhances performance from ground views by integrating complementary information from multiple branches, as demonstrated in the fourth column of Fig.~\ref{fig:ablation_street}. We also add two additional components based on the proposed initialization scheme, both of which result in further performance improvements. Ultimately, we achieve the best results by combining these three components.

\begin{table}[]
\caption{The ablation study of the components of CrossView-GS on NYC and SF scenes of UC-GS dataset~\cite{ucgs}.}
\label{tab:ablation}
\centering
\begin{tabular}{c|ccc|c|c}
\hline
\multirow{2}{*}{Variants} & \multicolumn{3}{c|}{All} & Aerial & Ground \\ 
                         & PSNR   & SSIM  & LPIPS  & PSNR  & PSNR  \\ \hline
Baseline                 & 26.464  & 0.771  & 0.302 & 27.998 & 24.929 \\
w/ Init.                & 26.814  & 0.780  & 0.292 & 28.592 & 25.037 \\
w/ Grad.                & 27.241  & 0.791  & 0.263 & 29.184 & 25.299 \\
w/ Supp.                & 27.174  & 0.792  & 0.272 & 28.971 & 25.377 \\
                         \hline
w/ Init. \& Grad.       & 27.457  & 0.797  & 0.257 & 29.505 & 25.409 \\
w/ Init. \& Supp.       & 27.333  & 0.796  & 0.266 & 29.242 & 25.425 \\
                         \hline
Ours                     & 27.829  & 0.809  & 0.241 & 29.844 & 25.814 \\ \hline
\end{tabular}
\end{table}

\subsection{Limitations}
\begin{figure}[h]
  \includegraphics[width=\linewidth]{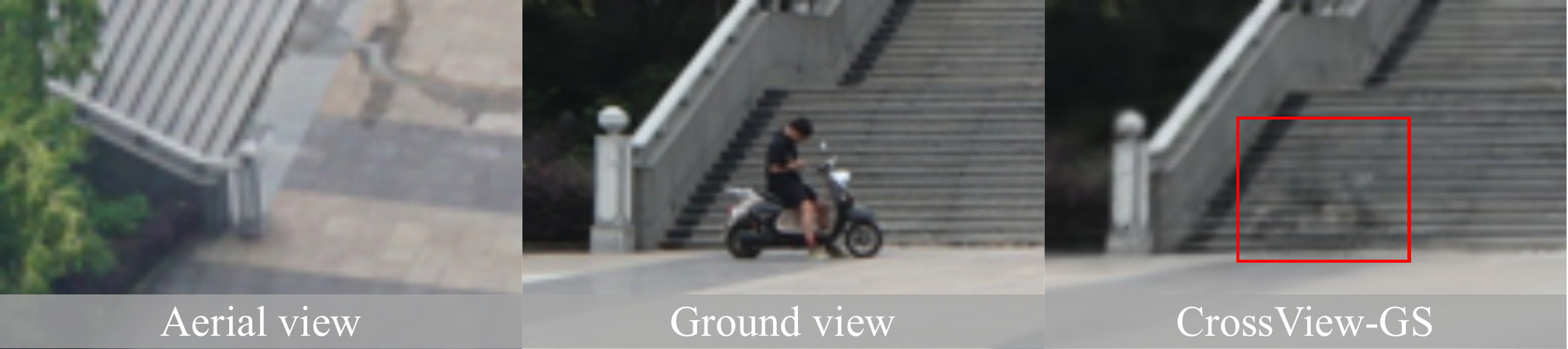}
  \caption{\textbf{Limitation}. The cross-view inconsistency caused by dynamic objects results in unnatural artifacts.}
  \label{fig:limitation}
\end{figure}

While our method effectively reconstructs large-scale scenes using cross-view data, it is not without limitations. Specifically, our method does not account for dynamic objects in the scene, such as vehicles and pedestrians. The presence of these objects in cross-view data introduces additional spatial ambiguity, as shown in Fig.~\ref{fig:limitation}. Although predicting masks for moving objects may offer a potential solution, its feasibility from aerial views remains uncertain.

\section{Conclusion}
We propose a novel method CrossView-GS, for large-scale scene reconstruction based on cross-view data. The multi-branch models is constructed as a prior by separately reconstructing different sets of views, which can effectively guide the optimization of cross-view Gaussian primitives and achieve high-quality reconstruction of large-scale scenes. Experimental evidence demonstrates that our method outperforms state-of-the-art methods. 

As the future work, we plan to incorporate the divide-and-conquer strategy to enable the reconstruction of extremely large-scale scenes using cross-view data.

\bibliographystyle{IEEEtran}
\bibliography{bare_jrnl_new_sample4}

\clearpage

\begin{figure*}[htbp]
  \includegraphics[width=\linewidth]{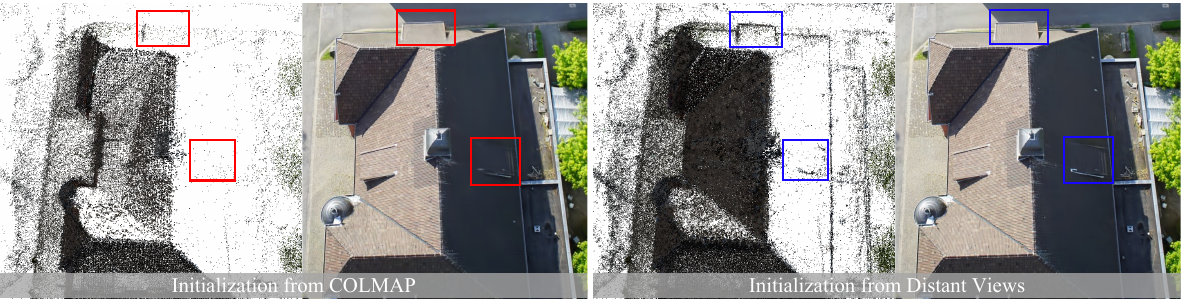}
  \caption{Point clouds and final reconstruction results by using different initializations.}
  \label{fig:init}
\end{figure*}

\clearpage

\appendix

\section*{A. Dataset Details} 
\begin{figure}[]
  \includegraphics[width=\linewidth]{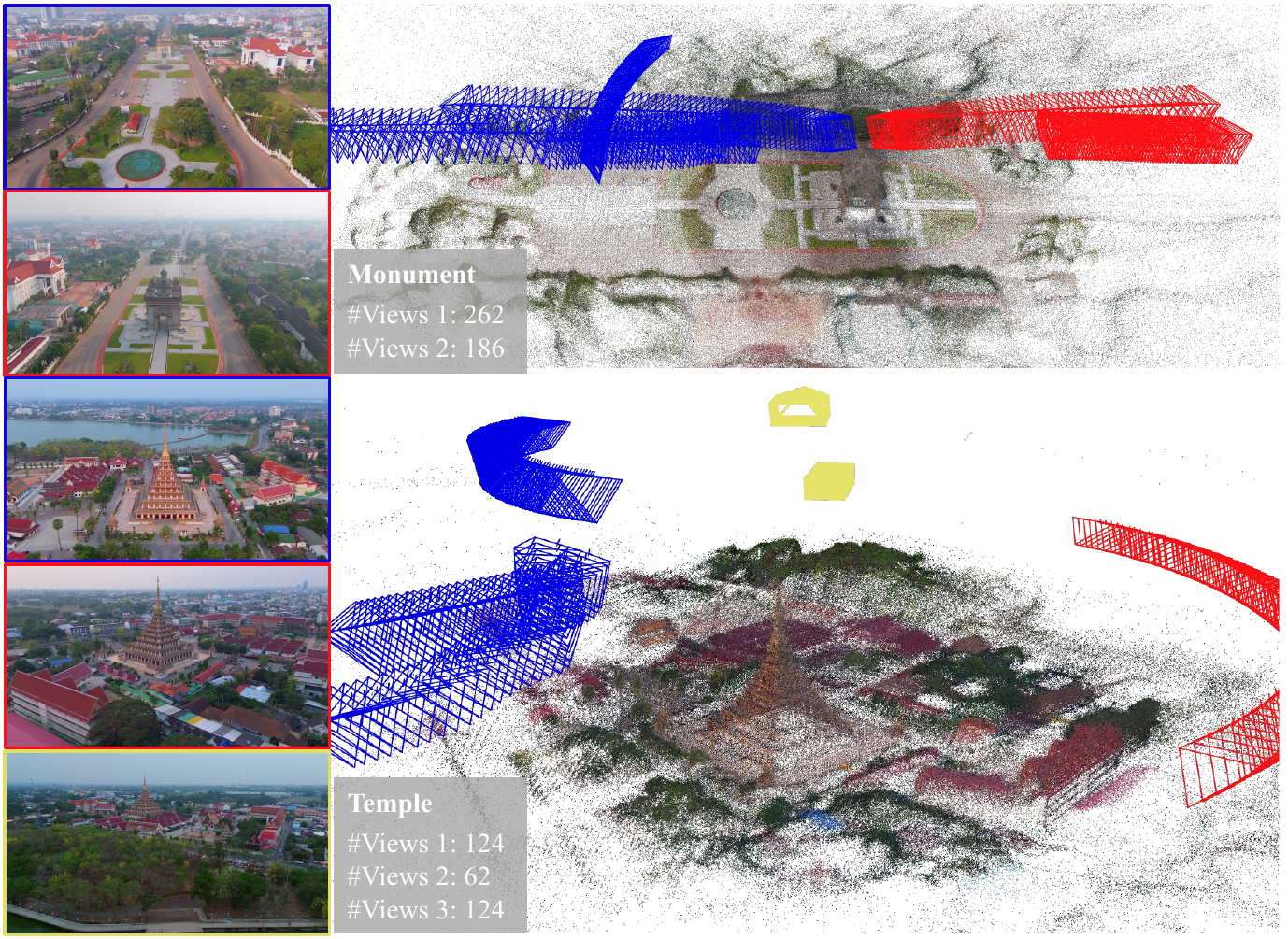}
  \caption{Visualization of Temple and Monument, as well as example views from different sets of views.}
  \label{fig:t_m}
\end{figure}

\textbf{Pure aerial views.} We have collected publicly available videos, casually captured at two landmark locations, Wat Nong Waeng Temple and Patuxai Monument, sourced from the Internet\footnote{https://www.youtube.com/@Laygtraveller}. Two datasets, named Temple and Monument, are constructed by applying simple trajectory-based clustering with several sets of views. For each video clip, we sample approximately 60 frames and select one frame out of every four to form the test set. Camera pose estimation is performed using MASt3R-SfM~\cite{mast3rsfm}, which demonstrates strong performance in cross-view registration. Visualizations of Temple and Monument datasets, along with example views from different sets highlighted in different colors, are shown in Fig.~\ref{fig:t_m}.

\textbf{Pure ground views.} In the MatrixCity dataset \cite{mc}, ground views are captured from five distinct directions: front, back, both sides, and top. Notable distinctions exist between the top views and the surrounding views. Specifically, in the Block\_A scene, we divide the ground views based on five different collection angles, distinguishing between front-back views and top views as two view sets. For the sake of clarity, we only use views from the most important region \textit{road\_down}, which includes 1,696 front-back views and 848 top views.

\section*{B. Analysis of unique Gaussian supplementation}
\begin{figure}[]
  \includegraphics[width=\linewidth]{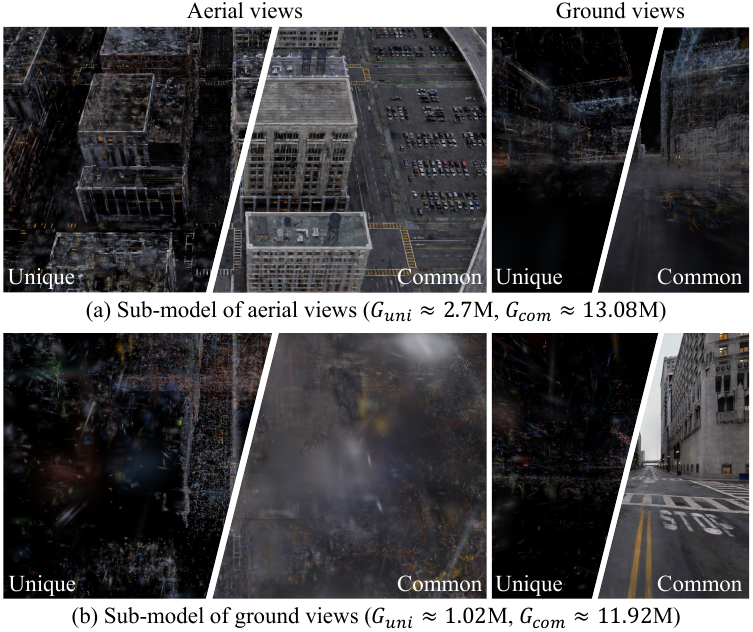}
  \caption{Visualization of unique Gaussian supplementation. We respectively present the rendering of the common and unique Gaussians of the different sub-models from different views in the MatrixCity dataset~\cite{mc}.}
\vspace{-0.2cm}
  \label{fig:unique_vis}
\end{figure}

In Fig.~\ref{fig:unique_vis}, we visualize the contributions of common and unique Gaussians across various sub-models and views. Generally, common Gaussians provide nearly complete rendering for their corresponding views, while unique Gaussians enhance fine details. As shown in the sub-model of aerial views in the first row, unique Gaussians supplement detailed information about roads and buildings from aerial views without introducing noise into ground views. Additionally, this visualization effectively illustrates the significant variations between cross-view data, underscoring the challenge of generalizing reconstructions from one type of view to another. 

In addition, we report the number of unique Gaussian primitives $G_{uni}$ and common Gaussian primitives $G_{com}$ of Fig.~\ref{fig:unique_vis}, with the unit in millions. In general, the number of unique Gaussians $G_{uni}$ only accounts for a small part of the total number of Gaussian primitives and is used to represent fine details.
In the Block\_small subset of the MatrixCity dataset, the cross-view model initially contains 13.69 million Gaussian primitives before supplementation. After supplementation and fine-tuning, this number increases to 16.41 million. In contrast, the variant that undergoes only fine-tuning contains 13.66 million Gaussian primitives and is unable to represent fine details effectively.

\end{document}